\documentclass{esannV2}
\usepackage{graphicx}
\usepackage[utf8]{inputenc}
\usepackage{amssymb,amsmath,array}
\usepackage{booktabs}
\usepackage{multirow}
\usepackage{amssymb}
\usepackage{url}
\usepackage{algorithm}
\usepackage[noend]{algpseudocode}
\usepackage[font=small,labelfont=bf]{caption}
\usepackage{caption}
\usepackage{subcaption}
\usepackage{hyperref}
\usepackage{float}
\usepackage{tablefootnote}
\begin{document}
\title{Continual Learning at the Edge: Real-Time Training on Smartphone Devices} 


\author{Lorenzo Pellegrini$^1$, Vincenzo Lomonaco$^2$, Gabriele Graffieti$^1$ and Davide Maltoni$^1$
%
%
\vspace{.3cm}\\
%
1- University of Bologna - Computer Science Department \\
Via Cesare Pavese, 50, 47521 Cesena - Italy \\
%
\vspace{.1cm}\\
2- University of Pisa - Computer Science Department \\
Largo B. Pontecorvo, 3, 56127, Pisa - Italy
}

\maketitle

\begin{abstract}
On-device training for personalized learning is a challenging research problem. Being able to quickly adapt deep prediction models at the edge is necessary to better suit personal user needs. However, adaptation on the edge poses some questions on both the efficiency and sustainability of the learning process and on the ability to work under shifting data distributions. Indeed, naively fine-tuning a prediction model only on the newly available data results in catastrophic forgetting, a sudden erasure of previously acquired knowledge. In this paper, we detail the implementation and deployment of a hybrid continual learning strategy (AR1*) on a native Android application for real-time on-device personalization without forgetting. Our benchmark, based on an extension of the CORe50 dataset, shows the efficiency and effectiveness of our solution.
\end{abstract}



\section{Introduction}

Efficiently learning at the edge is a challenging research problem for current machine learning research. In particular, the peace of advancements within the deep learning field has often been linked with a significant increase in computational and memory requirements. Deep models are often trained on remote servers and only later deployed with frozen learning capabilities on embedded devices. However, adapting prediction models at the edge is fundamental to preserve the private nature of users' data and to customize prediction models on the fly based on the specific user needs, even without an internet connection. On-device personalization could indeed provide more targeted and adaptive functionalities based on the user interaction and newly acquired data. However, on-device training is not only a matter of efficiency (i.e., memory and computation) but mostly about learning from shifting data distribution. Accumulating all the data seen over the lifetime of the deployed system and re-train the whole model from scratch on all the data becomes quickly impossible, especially with frequent real-time updates. On the other hand, just updating the prediction model using only the newly available data incurs in catastrophic forgetting \cite{MCCLOSKEY1989109}.

This paper focuses on the design and deployment of a hybrid continual learning strategy called \textit{AR1*} \cite{lomonaco2020a} on a native Android application for real-time on-device personalization without forgetting. AR1* has been already shown to be efficient and flexible on natural video sequences constituting small non-i.i.d. batches of data \cite{pellegrini2020}\cite{lomonaco2020a}. However, running that algorithm on edge devices is not straightforward as it poses several additional challenges. In particular, in this work we: i) summarize AR1* and discuss how it can be better parametrized in terms of Accuracy-Computation-Memory trade-off (Section \ref{sec:ar1} and \ref{sec:core}); ii) detail the implementation of the Android application with native Java and C++ code based on Caffe framework (Section \ref{sec:core}); iii) introduce an extension of CORe50 for the life-after-deployment experiments running on CPU-only edge devices and we assess AR1* performances w.r.t. other state-of-the-art strategies (Section \ref{sec:exps}); iv) propose a two phases consolidation to achieve both real-time fast update and off-line delayed optimization (Section \ref{sec:core}). The native Android application source-code, along with the Avalanche-based \cite{lomonaco2021} scripts used to reproduce the paper results are made available to further stimulate research in this area\footnote{\url{https://github.com/lrzpellegrini/CL-CORe-App}}.


\section{AR1: a Flexible Hybrid Strategy for Continual Learning}
\label{sec:ar1}

\begin{figure}
\centering
\begin{subfigure}{.5\textwidth}
  \centering
  \includegraphics[width=0.90\columnwidth]{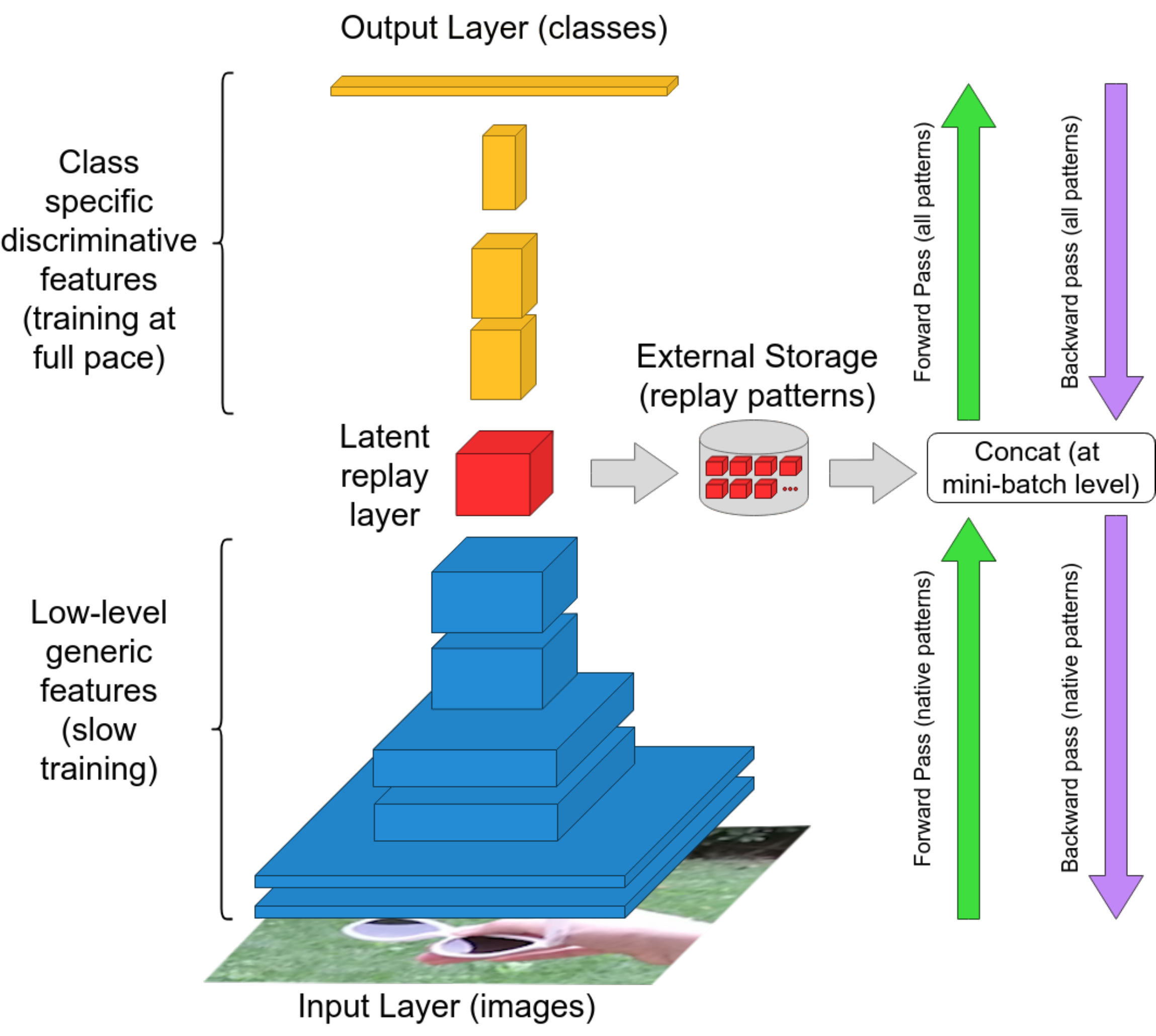}
  \label{fig:latent_replay}
\end{subfigure}%
\begin{subfigure}{.5\textwidth}
  \centering
  \includegraphics[width=0.90\textwidth]{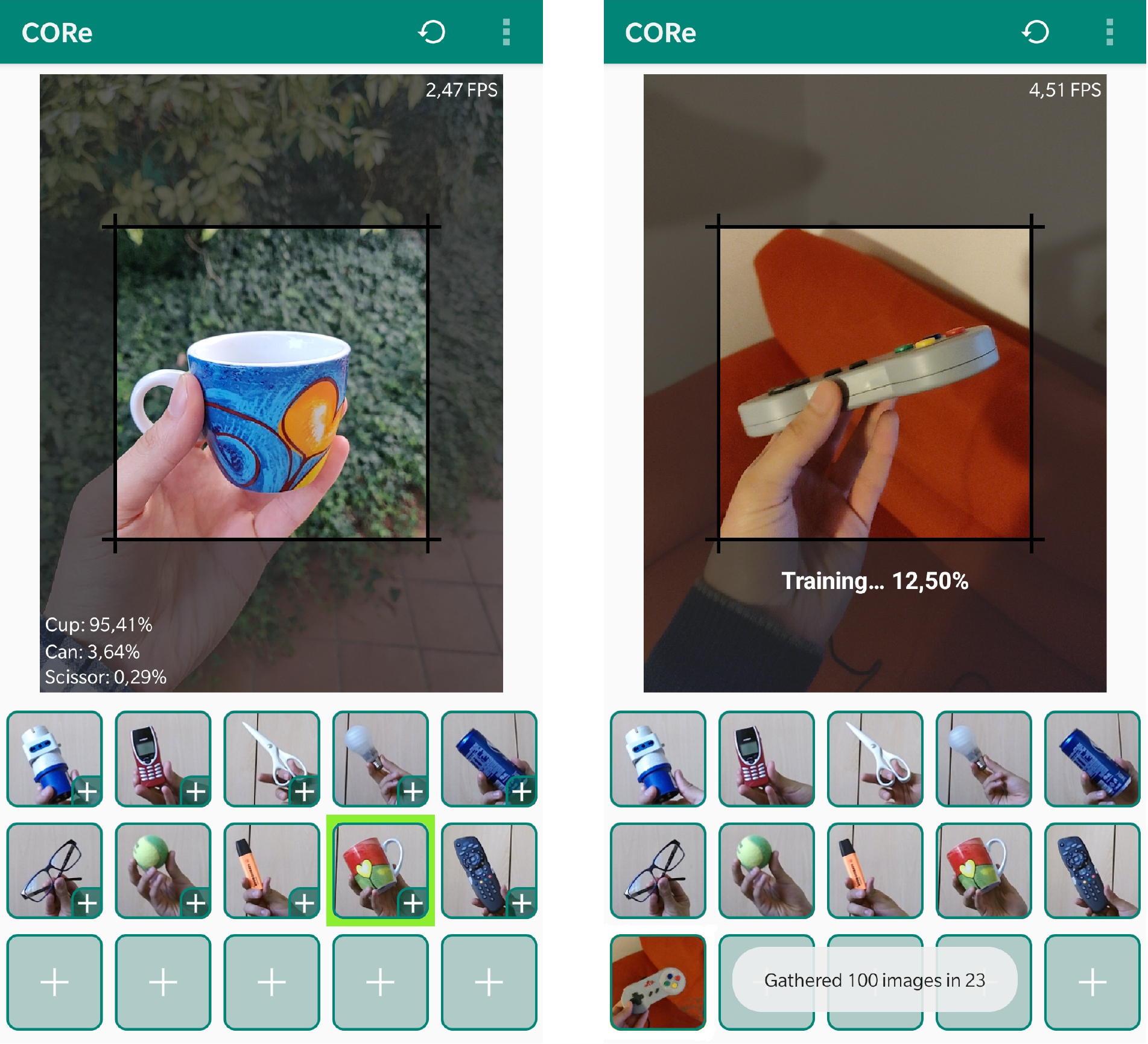}
  \label{fig:app_screenshot}
\end{subfigure}
\caption{Schema of AR1* with Latent Replay and the CORe Application User Interface.}
\label{fig:schema_and_app}
\end{figure}

In \cite{lomonaco2020a} two approaches denoted as CWR* and AR1* (hereafter \textit{CWR} and \textit{AR1} for brevity) were introduced and evaluated on CORe50 NICv2, a challenging continual learning setting where examples belonging to both known classes as well as completely new classes are encountered over time. CORe50 NICv2 is composed of 391 small non-i.i.d. batches, each consisting of 300 frames belonging to a short video of a single class.

In CWR the last fully connected layer is implemented as a double memory, and simple initialization and fusions steps are performed before and after each training batch to synchronize the two memories; this was shown to offer good protection against forgetting. However, after the first training batch, CWR freezes all the layers except the last one, thus losing the benefits of a continual adaptation of the underlying representation. AR1 extends CWR by enabling end-to-end continual training throughout the entire network by adopting a regularization approach to constrain the change of critical weights. 

While AR1 proved to be effective on CORe50 NICv2, the authors of \cite{pellegrini2020} proposed to extend it with a latent replay approach in order to close the gap w.r.t. the offline training upper bound. Latent replay (see Figure \ref{fig:schema_and_app}) denotes an approach where, instead of maintaining copies of input patterns in the external memory, activations volumes are stored at a given layer (denoted as \emph{latent replay layer}). When latent replay is implemented with mini-batch SGD training: \emph{(i)} in the forward step, a concatenation is performed at the replay layer (on the mini-batch dimension) to join patterns coming from the input layer with activations coming from the external storage; \emph{(ii)} the backward step is stopped just before the replay layer for the replay patterns.

\section{CORe: an Android Application Based on Caffe}
\label{sec:core}

The (C)ontinual (O)bject (Re)cognition application is an Android application running on common smartphones\footnote{CORe App video demo: \url{http://bit.ly/latent-replay}}. The goal of CORe is to show how continual learning can solve a practical computer vision problem. The target scenario resembles a realistic lifelong learning setup where the user tries to improve the object recognition capability of its knowledge model; the user can be a robot, a static vision system, or even a person wearing camera-equipped smart glasses. 

The application starts in inference mode: a continuous stream of frames is received from the rear camera and shown on screen. In inference mode, the application continually shows the top-3 predicted categories for the object depicted in the central part of the image (highlighted by a surrounding greyed out area). The application is packaged with a model pretrained on the CORe50 dataset, which consists of video sessions of objects commonly found in domestic or office environments. CORe50 contains video sessions of 50 different objects belonging to 10 categories \cite{lomonaco2017}. Considering that the goal of the application is a coarse-grained classification, the initial model here used was pretrained on the categories instead of the objects. The user can switch to training mode by either selecting an existing category or an empty slot. By selecting an existing category the user can incrementally improve the recognition capability of the model on that category, which can be useful if an object is misclassified. A new category can be added by selecting one of the five empty slots in the last row. This triggers the image gathering phase, during which a 20 seconds video is taken, for a total of 100 frames. The training phase is then carried out using the AR1 algorithm with latent replay; in the current version of the application latent replay takes place at the penultimate layer of the model (\textit{pool}), but alternative implementations are possible. In particular, a two stages consolidation can be envisaged where a quick (real-time) update takes place at the output layer and a slower (but more precise knowledge organization) is performed in background affecting deeper layers. This is also in line with biological learning where the hippocampus can make recent knowledge immediately accessible while consolidation in the cortex is carried out throughout sleep-awake cycles \cite{hayes2021replay}.
The application is equipped with a MobileNetV1-1.0 in 32-bit floating point format working with 128x128 input images \cite{howard2017mobilenets} and leverages a fixed-size replay buffer of just 500 patterns for which only latent representations are kept. The application uses the latent replay mechanism to drastically reduce the training time: latent activations of the new images are computed while the frames are gathered, so that at the end of acquisition the sole part of the model following the latent replay can be quickly trained with these precomputed activations. The performance/accuracy trade-off of this choice is discussed in detail in Section \ref{sec:exps}.
It should be noted that using the penultimate layer as the latent replay layer, the memory occupation of a replay buffer of 500 patterns is less than 2MByte.

A custom version of the Caffe framework is used for both the inference and training phases. Other popular deep learning libraries for mobile devices (i.e., TensorFlow-Lite, PyTorch-Mobile) still do not support training operations as they only package the operators strictly required for the forward pass. Moreover, to operate with these libraries, the model has to be properly adapted and frozen.
On the contrary, by compiling Caffe and its dependencies from scratch one can exploit complete deep learning functionalities. In Section \ref{sec:exps} we compare our apporach with a similar Android application based on TensorFlow-Lite proposed by \cite{demosthenous2021continual}, which uses an experimental external library to handle the training on the last fully connected layer. This library works atop of a TensorFlow-Lite model used as a fixed feature extractor.

\begin{figure}[t]
\centering
\begin{subfigure}{.5\textwidth}
  \centering
  \includegraphics[width=1.0\linewidth]{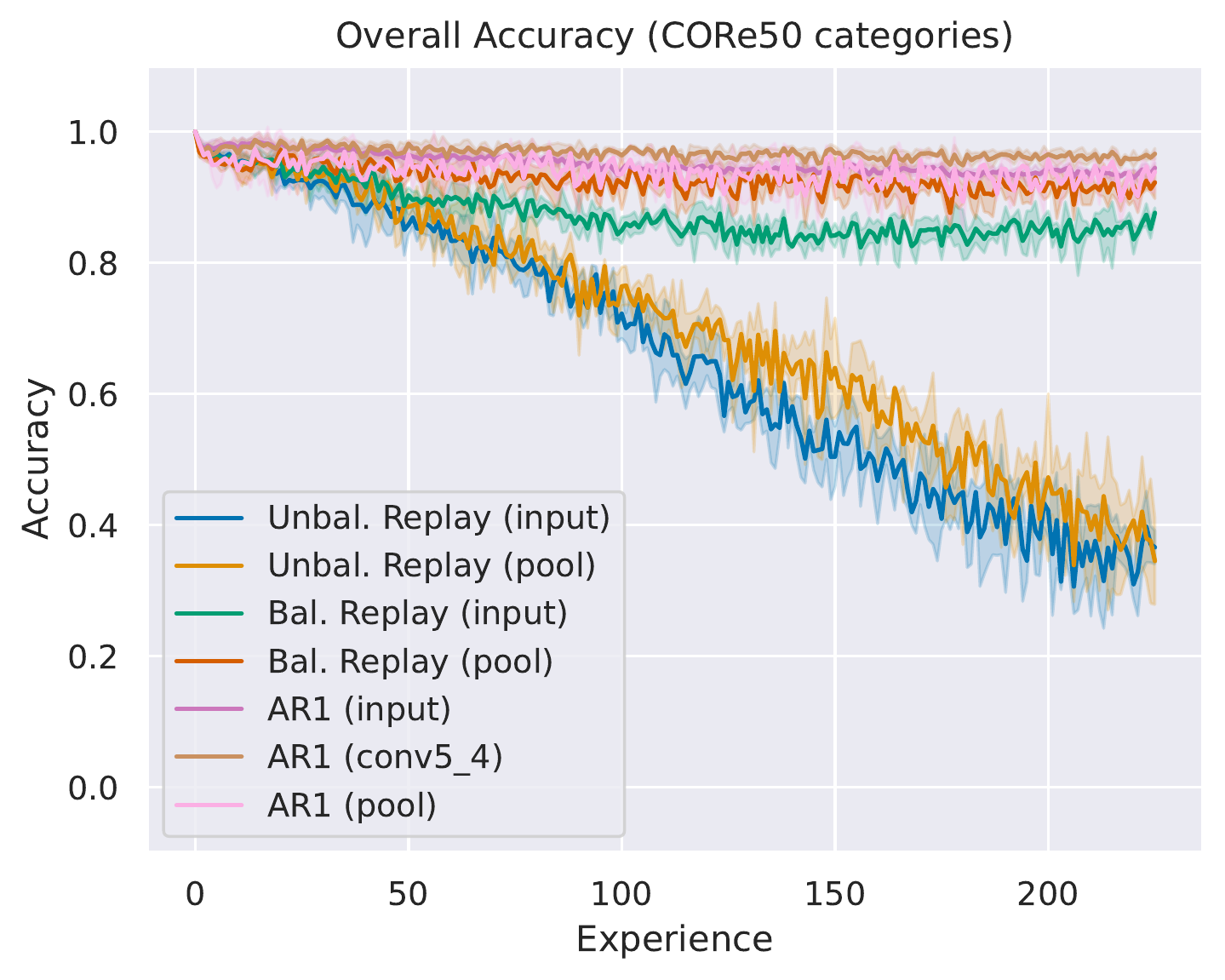}
  \label{fig:expCORe50}
\end{subfigure}%
\begin{subfigure}{.5\textwidth}
  \centering
  \includegraphics[width=1.0\linewidth]{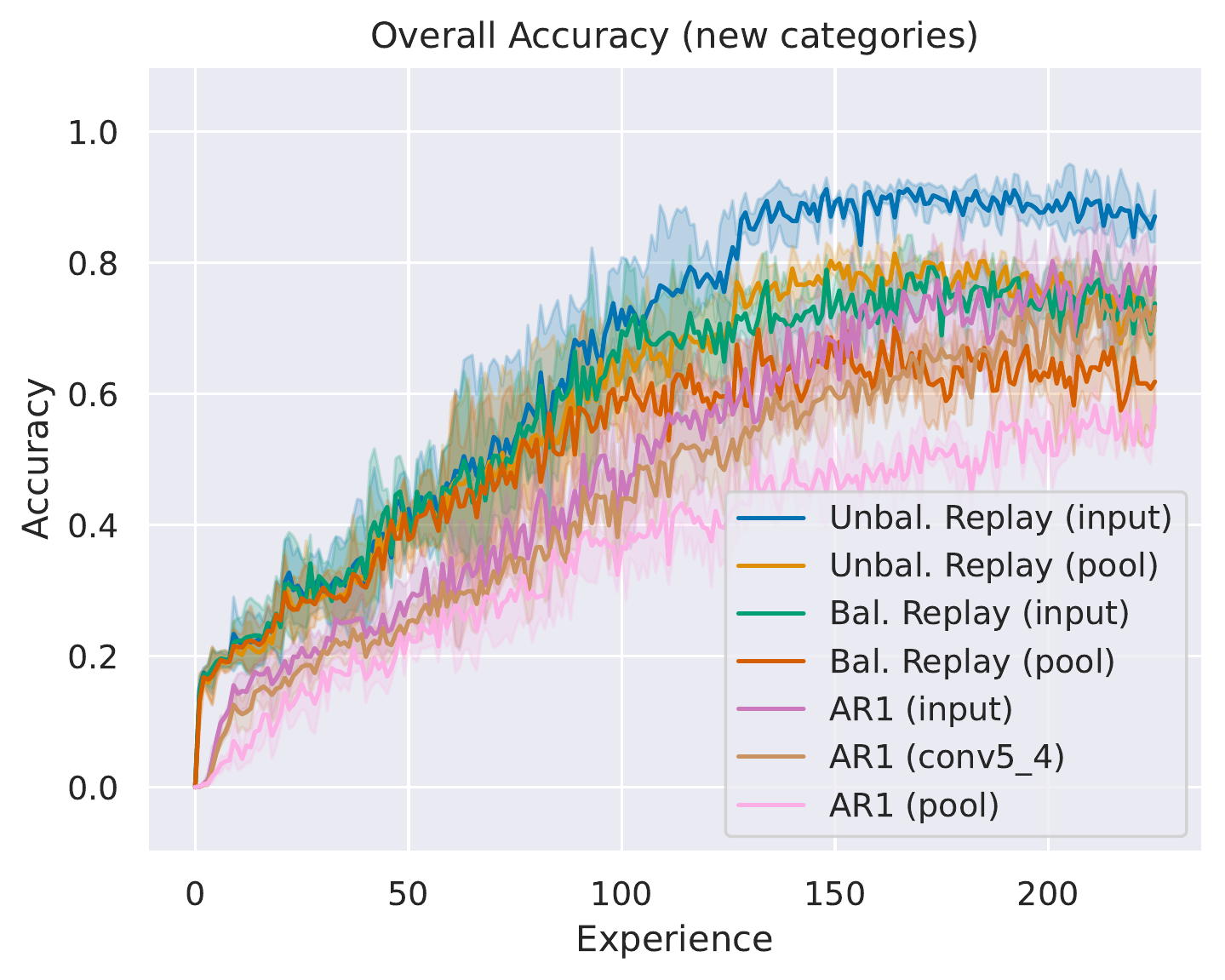}
  \label{fig:expNewCat}
\end{subfigure}
\vspace*{-5mm}
\caption{Test accuracy computed after each training experience for each strategy. Left: accuracy on the CORe50 categories; right: the accuracy on the new categories.}
\label{fig:experiments}
\end{figure}


\section{Experiments}
\label{sec:exps}

A proper benchmark must be defined in order to assess the ability of the CORe application to learn new categories of objects. In \cite{lomonaco2020a} the complex NICv2-391 benchmark is proposed where the system is required to continually learn from a stream of 391 batches (hereafter called \textit{experiences}). In that benchmark, each experience can either depict an already seen or a new object which makes it very akin to the use case covered by the CORe application. NICv2-391 cannot be used here because CORe50 already serves as the pretraining dataset used to populate the initial 10 categories. However, following the idea behind NICv2, here we propose an extension of the benchmark where new categories are encountered over time. Therefore, we collected an extended dataset containing 25 objects grouped in 5 new categories; 12 sessions (both indoor and outdoor) are available for each object: 9 are used for training while 3 are left for the test set. Therefore, the final benchmark consists of 225 experiences and it is modeled in the same way as NICv2 by following the generation procedure introduced in \cite{lomonaco2020a}. The dataset was captured by using different smartphones thus obtaining videos whose quality, field-of-view and stability are different from those in CORe50. Our experiments are aimed at measuring the degree of forgetting on the original CORe50 categories in the complex scenario in which no video sessions depicting objects of these categories are seen again.

\begin{table}[ht]
\centering
\caption{Final accuracy and on-device training time (in seconds). The \textit{pool} variant allows for shorter training phases while sacrificing the accuracy on new categories. The \textit{conv5\_4} variant is a good trade-off between \textit{pool} and \textit{input}.}
\label{table:timesTable}
\begin{tabular}{lr|rrr}
                                                                                              & AR1 variant:                                                                                                                                                                                               & \multicolumn{1}{c}{\textbf{Pool}} & \multicolumn{1}{c}{\textbf{Conv5\_4}} & \multicolumn{1}{c}{\textbf{Input}}  \\ 
\hline\hline
\multirow{2}{*}{Final accuracy}                                                               & Initial categories                                                                                                                                                                                                   & \textbf{97.54\%}                  & 97.16\%                               & 93.62\%                             \\
                                                                                              & New categories                                                                                                                                                                                                       & 65.33\%                           & 73.39\%                               & \textbf{79.07\%}                    \\ 
\hline
\multirow{5}{*}{\begin{tabular}[c]{@{}l@{}}Training time for \\a new experience on\\100 new patterns,\\ 8 epochs\end{tabular}} & Feature extraction\tablefootnote{The feature extraction step is executed concurrently with the image gathering phase and does not affect the user experience.} & 8.93                              & 6.70                                  & \multicolumn{1}{c}{-}               \\
                                                                                              & Forward pass                                                                                                                                                                                                         & 0.14                              & 125.05                                & 436.33                              \\
                                                                                              & Backward pass                                                                                                                                                                                                        & 0.04                              & 4.82                                  & 4.81                                \\
                                                                                              & Weights update                                                                                                                                                                                                       & 0                                 & 37.21                                 & 58.88                               \\ 
\cline{2-5}
                                                                                              & Overall                                                                                                                                                                                              & 0.18                              & 167.08                                & 500.02                             
\end{tabular}
\end{table}

In the proposed experiments the AR1 strategy is compared against two pure-replay strategies namely “class balanced” and “unbalanced”. In order to allow for a fair comparison, the same setup used in the application is considered: the fixed-size replay buffer (500 patterns) is initially filled with instances from the CORe50 dataset by selecting the same number of instances for each class. In all the compared strategies, the instances to be inserted and replaced are chosen randomly. In particular, the “class balanced” approach balances the amount of instances contained in the replay buffer across categories while the “unbalanced” strategy does not employ any kind of balancing mechanism and replaces a fixed amount of patterns (10) after each training experience. For the replay strategies, the result of their \textit{\textit{pool}} variant is reported, in which the model layers below the last fully connected layer are kept frozen, thus using it as a feature extractor. For AR1, the \textit{\textit{pool}} and \textit{\textit{conv5\_4}} variants identify the chosen latent replay layer. It is worth noting that the “unbalanced (pool)” strategy is the same one employed in the application described in \cite{demosthenous2021continual}, which serves here as a comparison with the proposed AR1 strategy. For all strategies, the \textit{\textit{input}} variant identifies the replay from the input layer.

The experimental results reported in Figure~\ref{fig:experiments} and Table~\ref{table:timesTable} show that AR1 provides a good trade-off since it is able to learn the new categories without forgetting the old ones. The unbalanced replay mechanism shows good learning capabilities on the new categories but is not able to avoid forgetting on the CORe50 ones, on which the accuracy steadily decreases. Nevertheless, this minimal replay mechanism is still able to prevent an abrupt accuracy drop. The \textit{pool} variant of the class balanced replay is able to retain the knowledge of the CORe50 categories but it does not reach the same accuracy performance of AR1 on the new categories.

The training phase of the chosen continual learning algorithm has to be completed in a certain time span based on the target user experience. The AR1 strategy allows for a fine-grained trade-off choice between accuracy and performance based on the selected latent replay layer. 
Table~\ref{table:timesTable} shows a comparison for these different choices. 
Using \textit{conv5\_4/dw} as the latent replay layer allows for near-optimal knowledge conservation on the original CORe categories while demonstrating a good training trend on the new ones.
A further optimization could be achieved by combining \textit{pool} and \textit{conv5\_4} according to the two steps consolidations discussed in Section \ref{sec:core}.


\section{Conclusion and Future Work}
\label{sec:conclusion}

Learning continually at the edge 
may open the door to a number of privacy-preserving and personalized AI systems. However, on-device training is subject to many real-world constraints, strict computational and memory limitations. In this paper, we showed that a hybrid continual learning strategy, \textit{AR1}, can provide an efficient and effective approach for sustainable on-device personalization while controlling forgetting on previously acquired knowledge.



\begin{footnotesize}

\bibliographystyle{unsrtnamed}
\bibliography{main.bib}


\end{footnotesize}


\end{document}